# Large language models can replicate cross-cultural differences in personality

**Paweł Niszczota [a,*], Mateusz Janczak [a], Michał Misiak [b,c]**

[a] Humans & AI Laboratory (HAI Lab), Institute of International Business and Economics, Poznań University of Economics and Business, Poznań, Poland

[b] IDN Being Human, Institute of Psychology, University of Wrocław, Wrocław, Poland

[c] School of Anthropology & Museum Ethnography, University of Oxford, Oxford, United Kingdom

[*] Corresponding author: Paweł Niszczota, Poznań University of Economics and Business, al. Niepodległości 10, 61-875 Poznań, Poland, pawel.niszczota@ue.poznan.pl

**Data**

Data and the pre-registration document are available at: https://osf.io/76g2u/.

**Declaration of Competing Interest**

The authors declare that they have no known competing financial interests or personal relationships that could have appeared to influence the work reported in this paper.

**Ethical approval**

The study did not involve any human subjects and did not require ethical approval.

**Financial support**

This research was supported by grant 2021/42/E/HS4/00289 from the National Science Centre, Poland.

**Acknowledgments**

We'd like to thank Dirk Wulff for valuable comments on the manuscript.

**CRediT authorship contribution statement**

**Paweł Niszczota**: Conceptualization; Methodology; Software; Validation; Formal analysis; Investigation; Resources; Data Curation; Writing – Original Draft; Writing – Review & Editing; Visualization; Supervision; Project administration; Funding acquisition

**Mateusz Janczak**: Conceptualization; Formal analysis; Investigation; Writing – Review & Editing

**Michał Misiak**: Formal analysis; Investigation; Methodology; Writing – Review & Editing

**Published in *Journal of Research in Personality* – this is the Author Accepted Manuscript**

For final version (version of record) see:
Niszczota, P., Janczak, M., & Misiak, M. (2025). Large language models can replicate cross-cultural differences in personality. Journal of Research in Personality, 115, 104584. https://doi.org/10.1016/j.jrp.2025.104584

# Large language models can replicate cross-cultural differences in personality

## Abstract

We use a large-scale experiment (*N*=8000) to determine whether GPT-4 can replicate cross-cultural differences in the Big Five, measured using the Ten-Item Personality Inventory. We used the US and South Korea as the cultural pair, given that prior research suggests substantial personality differences between people from these two countries. We manipulated the target of the simulation (US vs. Korean), the language of the inventory (English vs. Korean), and the language model (GPT-4 vs. GPT-3.5). Our results show that GPT-4 replicated the cross-cultural differences for each factor. However, mean ratings had an upward bias and exhibited lower variation than in the human samples, as well as lower structural validity. We provide preliminary evidence that LLMs can aid cross-cultural researchers and practitioners.

## 1. Introduction

Even though large language models (LLMs) caught the attention of the public a while back (GPT-3, 2020), it was the introduction of ChatGPT that ignited a boom in interest in such models. This includes both laypeople and academics, including personality researchers (e.g., Jiang et al., 2023; Serapio-García et al., 2023). Some researchers (e.g., Horton, 2023) have proposed that LLMs could simulate human behavior. Recent advancements in Generative Pre-trained Transformer (GPT) models have revealed emerging capabilities (e.g., Bubeck et al., 2023), and perhaps simulating some patterns of human behavior is one of them. While LLMs should not be used as a replacement for human participants, they could be used as a sandbox for cross-cultural researchers and practitioners. These models could be used to conduct preliminary tests or to further validate findings obtained from human participants (Dillion et al., 2023).

In our study, we wanted to test whether LLMs are capable of mimicking real-world cross-cultural differences in personality. We used the Big Five model of personality (Gosling et al., 2003; Piedmont & Chae, 1997). The development of the Big Five was based on the lexical hypothesis, which posits that significant aspects of personality are reflected in vocabulary (Goldberg, 1993). Cutler and Condon (2023) propose that LLMs may resemble and potentially enhance methods used in early lexical studies on personality. They can assist in identifying personality descriptors, collecting data, and analyzing relationships between words, potentially offering new insights into personality structure. Contemporary psychologists often use traditional questionnaires for measuring personality traits and cross-cultural differences. We do not know, however, whether LLMs are able to mimic human responses to these kinds of measures through text generation (Hussain et al., 2024).

To test GPT's ability to simulate cross-cultural differences in personality, we chose the United States and South Korea as targets for this simulation. We have done this for two reasons. First, studies point to the existence of substantial differences in the mean scores of people from these two countries on the Big Five (Kajonius & Giolla, 2017; Piedmont & Chae, 1997; Yoon et al., 2002). Yoon and colleagues (2002) argue – while discussing their findings and the findings from earlier research – that it is unlikely that these differences are due to the result of artifacts introduced while translating the inventories used, and instead, they reflect actual cultural differences or different rating strategies used by members of these cultures.

Secondly, we used Korean as the alternative language due to its difference from the English language, on which GPT was mostly trained: notable differences were the basic sentence word order and the usage of different alphabets. In Table 1A, we present how mean scores differ in the original, US version of the Ten-Item Personality Inventory (TIPI; Gosling et al., 2003) and the Korean version of TIPI, i.e., TIPI-K, developed by Ha et al. (2013). Studies that directly compared US Americans against South Koreans using the Big Five taxonomy (see Tables S1-2) largely point to the same differences (Kajonius & Giolla, 2017; Piedmont & Chae, 1997).

While we are not the first to use LLMs to simulate human behavior (e.g., Aher et al., 2023; Horton, 2023; Serapio-García et al., 2023), we are the first to investigate whether LLMs can replicate cross-cultural personality differences. Let us consider the most intuitive mechanism that would allow LLMs to replicate cross-cultural personality differences, namely that LLMs obtain such an ability through extensive training on human-generated text, including discussions about personality and behavior (Bubeck et al., 2023; Hussain et al., 2024). This could enable LLMs to generate culturally-appropriate responses reflecting typical personality traits. LLMs' capacity to mirror personality expression across cultures would demonstrate progress in modeling human cognition and social behaviors (Dillion et al., 2023; Salah et al., 2023).

Cross-cultural studies often rely on adaptations of inventories, typically requiring the use of languages different from the original inventory's design. The difference in languages used is a potential problem for LLMs, as they have largely been trained on English text. However, some evidence suggests that GPT models perform well in languages other than English (Rathje et al., 2023). To test the current capabilities of the most advanced LLMs, we analyzed either the original languages of the inventories or held the language of the inventory constant. In the latter case, we varied not the language of the inventory, but the origin of the simulated target.

## 2. Hypotheses

In our main hypothesis, we posit that GPT-4 can replicate cross-cultural differences in Big Five personality factors. We focus on TIPI, which is a popular short inventory measuring the Big Five. Given that GPT is expected to respond differently depending on who is the person it is simulating, we used a system message (pre-prompt), instructing GPT to play the role of an adult from either USA or South Korea.

We wanted to answer two additional questions. Firstly, we asked whether GPT can replicate cross-cultural differences when using just one inventory (language), and not two. If personality differences between US Americans and South Koreans reflect actual cultural differences and are not artifacts, as argued, prompting an LLM in the same language should reproduce the expected personality differences. Secondly, we aimed to assess whether the development of GPT models shows progress in their potential ability to simulate human responses. We compared the performance of GPT-4 against the earlier generation, GPT-3.5.

## 3. Methodology

The study was conducted in August 2023 via API (programmatic) access to GPT. We used the following system message: "You are playing the role of an adult from [the United States/South Korea]". To elicit responses that could be efficiently analyzed, we modified the text used in the inventories, so that answers to each item were provided in square brackets (see Supplemental Materials for details). We used the least deterministic temperature setting (*temperature* = 2) to elicit the greatest level of variation in the answers.

The study had a 2 (target: from the United States vs. from South Korea) × 2 (inventory and language: TIPI in English vs. TIPI-K in Korean) × 2 (model: GPT-4 vs. GPT-3.5) design. As we wanted to collect 1000 observations per condition, we collected 8000 observations overall.

The main pre-registered test concerned the ability of GPT-4 to replicate differences in Big Five scores between US Americans and South Koreans. Given that this meant conducting five tests, to account for multiple tests, we used $\alpha = .05/5 = .01$ as the Type-I error rate. This gave us 95% power to detect an effect of $d = 0.189$, and 80% power to detect an effect of $d = 0.153$. Here we make the charitable assumption that observations generated by LLMs are independent – we discuss this issue further in the Supplemental Materials.

The study was pre-registered at https://aspredicted.org/zy8s-j37r. Data, code, and materials are available at https://osf.io/76g2u/.

## 4. Results

### 4.1. Structural validity and other properties of output

TIPI was specifically designed as a short measure of the Big Five personality dimensions, prioritizing content validity over traditional psychometric criteria. This design choice involves an inherent trade-off: while TIPI offers brevity and broad content coverage, its structure (two items per factor) precludes conventional measurement invariance analyses. This deliberate heterogeneity in item content, while enhancing construct representation, results in lower internal consistency estimates (Thørrisen & Sadeghi, 2023). To provide insights into the psychometric properties of generated TIPI and TIPI-K, we conducted a series of internal consistency analyses, including Cronbach's αs and Guttman's $\lambda_6$s (Table S3). Regarding αs, the most significant difference was observed for Emotional Stability, which had an α of .73 in the original study, but only .19 (TIPI) and .16 (TIPI-K) in the simulated data. To further test the internal consistency of GPT-4 generated data, we used the Spearman-Brown formula, which is recommended for two-item factors (Eisinga et al., 2013). These results were similar to those provided by classical internal consistency tests (Table S4). Our findings mirror the findings of a methodological review that demonstrated Extraversion has the highest internal consistency and Agreeableness the lowest (Thørrisen & Sadeghi, 2023).

### 4.2. Prompting in inventory-target congruent pairs

Our main hypothesis concerns the ability of GPT-4 to replicate cross-cultural differences in personality. As noted earlier, personality differences can be the result of actual differences between cultures or artifacts introduced during the adaptation of a personality inventory for a different culture. To stay the most comparable to the differences exhibited in TIPI and TIPI-K, we prompted GPT in English and Korean, respectively.

GPT-4 replicated all of the differences that were present in the original TIPI and TIPI-K studies: Korean targets in TIPI-K had lower scores on each of the Big Five factors (when using Emotional Stability, instead of Neuroticism). See Table 1A for test results and Figure S1 for an illustration of how the scores were distributed.

### 4.3. Prompting holding inventory constant

If cross-cultural differences in personality are substantial, then a practical alternative would be to prompt LLMs in just one language. As LLMs are usually trained mostly on text in English and inventories are also constructed in English, it seems plausible to simply construct prompts in this language.

Results – presented in Table 1B – show that GPT replicated almost all of the expected patterns (South Koreans obtaining lower scores than US Americans) both when it was prompted in English (using TIPI) or Korean (using TIPI-K). An exception was Agreeableness for TIPI-K, where the difference had the opposite sign than expected.

### 4.4. Accuracy of scores

LLMs can replicate differences in scores between cultures without accurately reflecting actual mean levels. For example, an LLM can detect that Koreans on average obtain lower scores on a personality trait than Americans, yet the simulated values could be offset (e.g., inflated) in both cases. Therefore, we also compared how accurately GPT reflected mean personality scores, working under the assumption that TIPI in both the original (US) version and the adapted (Korean) version accurately captured variation in the Big Five.

#### *Inventory-target congruent pairs*

Raw differences suggest that there was a general upward bias in responses, for both the TIPI-US pair ($M$ = 1.11) and TIPI-K-Korean pair ($M$ = 1.70), being significantly lower in the former ($d$ = –0.50 [–0.54, –0.46]). We speculate on the potential reason behind this tendency in the discussion.

**Table 1**

*Actual and simulated scores for TIPI-US and TIPI-K-Korean pairs (Panel A), and simulated scores for when using the same inventory (Panel B)*

**Panel A. Comparison of actual and simulated scores: US target in TIPI (English) vs South Korean in TIPI-K (Korean)**

| Factor | Actual | | | Simulated (GPT-4) | | |
|---|---|---|---|---|---|---|
| | US/TIPI | South Korean/TIPI-K | Cohen's *d* | US/TIPI | South Korean/TIPI-K | Cohen's *d* |
| Extraversion | 8.88 (2.90) | 8.47 (2.67) | 0.15 | 10.04 (0.99) | 9.60 (1.55) | 0.34 [0.25, 0.43] |
| Agreeableness | 10.46 (2.22) | 9.50 (2.07) | 0.48 | 11.99 (0.74) | 11.25 (0.94) | 0.88 [0.78, 0.97] |
| Conscientiousness | 10.80 (2.64) | 9.42 (2.31) | 0.56 | 12.57 (0.75) | 11.57 (0.98) | 1.10 [1.00, 1.20] |
| Emotional Stability | 9.66 (2.84) | 8.36 (2.43) | 0.49 | 10.10 (0.81) | 9.56 (1.20) | 0.52 [0.43, 0.61] |
| Openness to Experiences | 10.76 (2.14) | 8.46 (2.51) | 0.99 | 11.43 (1.05) | 10.71 (1.35) | 0.59 [0.50, 0.68] |

**Panel B. Simulated scores for both targets using same inventory, i.e. TIPI (English) vs TIPI-K (Korean)**

| Factor | TIPI | | | TIPI-K | | |
|---|---|---|---|---|---|---|
| | US | South Korean | Cohen’s *d* | US | South Korean | Cohen’s *d* |
| Extraversion | 10.04 (0.99) | 8.90 (1.00) | 1.10 [1.10, 1.20] | 10.66 (1.42) | 9.60 (1.55) | 0.72 [0.63, 0.81] |
| Agreeableness | 11.99 (0.74) | 11.70 (0.72) | 0.39 [0.30, 0.48] | 11.13 (0.95) | 11.25 (0.94) | –0.12 [–0.21, –0.03] |
| Conscientiousness | 12.57 (0.75) | 12.43 (0.79) | 0.18 [0.09, 0.27] | 11.71 (1.09) | 11.57 (0.98) | 0.13 [0.04, 0.22] |
| Emotional Stability | 10.10 (0.81) | 9.99 (0.84) | 0.12 [0.04, 0.21] | 9.80 (1.13) | 9.56 (1.20) | 0.21 [0.12, 0.29] |
| Openness to Experiences | 11.43 (1.05) | 10.96 (0.95) | 0.46 [0.37, 0.55] | 11.19 (1.32) | 10.71 (1.35) | 0.36 [0.28, 0.45] |

*Note*. This table presents means and standard deviations (brackets), and effect sizes [with 95% *CI*s]. All pre-registered *t*-tests that compare the simulated scores for the US target using TIPI and South Korean target using TIPI-K (Panel A) indicate that the differences are statistically significant at $p < .0001$.

**Figure 1**

*Accuracy of GPT-4 across the Big Five*

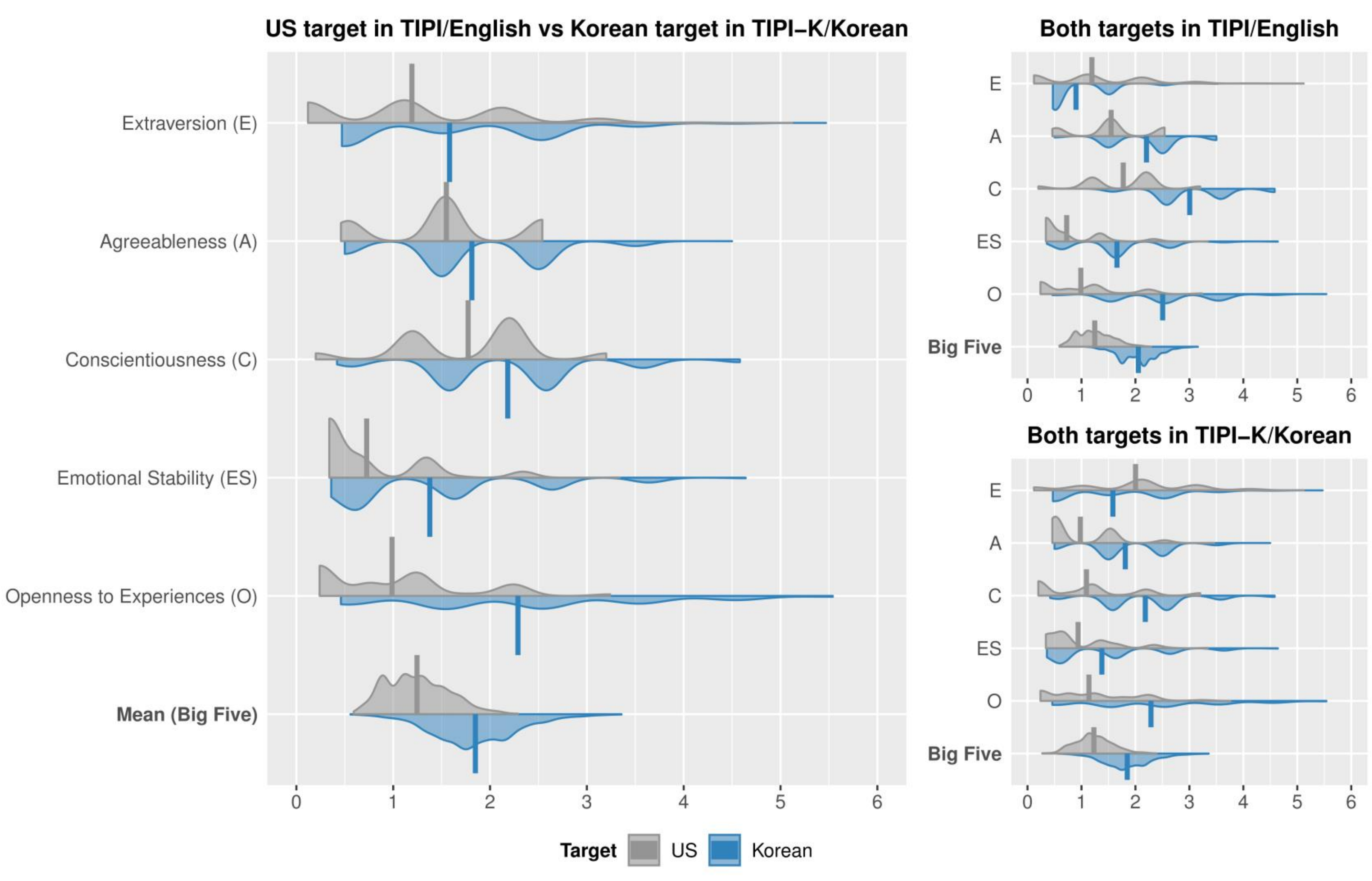

*Note*. Bars represent means. Accuracy is the absolute value of the difference between simulated and actual scores.

The absolute difference between simulated and original scores (or the Mean Absolute Error) – shown in Figure 1 – pointed to the accuracy being greater for the TIPI-US pair ($M$ = 1.25) than the TIPI-K-Korean pair ($M$ = 1.85, $d$ = –0.62 [–0.66, –0.58].

There was variation across the accuracy of simulated scores, with the lowest being for Emotional Stability in the TIPI-US pair ($M$ = 0.73) and the highest being for Openness to Experience in the TIPI-K-Korean pair ($M$ = 2.29).

***Holding inventory constant***

Instead of varying both the inventory (TIPI vs. TIPI-K) and the target (US vs. Korean), we can compare the accuracy of scores when only manipulating the target (a person from the United States vs. a person from South Korea). In our case, we can either use TIPI or TIPI-K as the benchmark inventory. As shown in Figure 1, accuracy when using TIPI (i.e., in English), was better for US than South Korean targets, $d$ = –0.62 [–0.66, –0.58]. It was also better for US than South Korean targets when using TIPI-K (i.e., in Korean), $d$ = –0.80 [–0.85, –0.77].

## 5. Discussion

Large language models can potentially encapsulate human behavior across various hypothetical scenarios, replicating numerous prominent findings in psychological science (Horton, 2023). This includes the structure of personality traits (Cutler & Condon, 2023). In our study, we aimed to advance this field of research by testing whether LLMs could replicate prominent cross-cultural differences, such as those observed between US Americans and South Koreans on the Big Five personality traits (Kajonius & Giolla, 2017; Piedmont & Chae, 1997; Yoon et al., 2002).

Using the Ten-Item Personality Inventory (Gosling et al., 2003) to assess personality based on the Five-Factor Model, our results show that GPT-4 can replicate all of the differences previously observed when comparing Americans and South Koreans on the Big Five factors. Specifically, our simulated South Korean responses showed lower scores on Extraversion, Agreeableness, Conscientiousness, and Openness to Experiences and higher scores on Neuroticism. This provides initial support for the use of LLM for *in silico* experiments, where targets of simulations are meant to represent people from different cultures.

LLMs cannot replicate all of the characteristics of people that were sampled in the US and South Korean TIPI studies. There was an upward bias in scores, variation was smaller than in the original samples, and the factors had substantially lower structural validity. Compared to GPT-4, output from GPT-3.5 exhibited extremely poor structural validity. Probably due to this, it did not entirely replicate the expected cross-cultural differences (see Figure S1 in the Supplemental Materials).

While GPT-4 replicated differences for all five factors, the difference between actual and simulated scores was not constant. To illustrate, the difference in Openness to Experience between TIPI and TIPI-K scores is a substantial 0.99 in real-world data. In contrast, for simulated scores the difference is only 0.59 (40% less). Differences between simulated scores of US and Korean targets were the greatest for Agreeableness, even though prior studies do not suggest that this is the dimension in which people from the two investigated cultures differ the most.

***Other potential mechanisms***

The introduction outlined a default mechanism for LLMs' performance: uncovering actual behavioral, cognitive, and affective tendencies of people from training data. However, two other potential, non-mutually exclusive mechanisms may be driving our results. One possibility is that when playing the role of people from different countries, GPT is relying on stereotypes, and simply trying to mimic them when instructed to play a role of a person from a given country. However, research on national stereotypes shows that these are often inaccurate (Jussim et al., 2016), and thus would not help replicate real-world patterns.

A second possibility is that fine-tuning by humans through Reinforcement Learning from Human Feedback (RLHF) – intended to make LLMs more user-friendly and reduce toxic content – might be responsible for the findings. However, the simulated differences are still substantial and opposite to the direction expected from aggressive socially desirable adjustments. Nonetheless, RLHF could be at play. It is plausible that the upward bias in scores present both for Americans and South Koreans was the result of RLHF inflating the scores of simulated persons regardless of origin. RLHF could also be at least partially responsible for the variation in LLM's accuracy across factors, that we mentioned at the end of the last subsection.

We offer a more detailed discussion on the potential roles of stereotypes and RLHF in the Supplementary Materials. Note that this discussion is largely speculative: more research is needed to learn "what is happening under the hood" of LLMs. This might require the use of more open (and thus controllable) LLMs and alternative methodological approaches (e.g., Cutler & Condon, 2023).

***Limitations on Generality***

In our study we used one personality model (the Five-Factor model), one personality inventory (TIPI), one cross-cultural pair (United States-South Korea), and an LLM originating from one source (GPT from OpenAI). Our findings should be validated using different personality models and inventories, different cross-cultural pairs, and different LLMs. Moreover, we have prompted the LLM to play the role of an adult, but have not specified any other personality-relevant requirements for the simulated person, such as their gender.

***Future directions***

Apart from encouraging researchers to address the limitations highlighted above, we would like to point to a particularly interesting future direction. Given that there are prominent gender differences in personality, and that the stereotypes about these gender differences differ across cultures (Giolla & Kajonius, 2019), a fruitful avenue would be to test whether LLMs could replicate these differences cross-culturally, using different languages. This could further test the possibility that LLMs are informed by stereotypes to generate their responses. This could be achieved by using something akin to our text generation approach, or rely on an analysis of word embeddings in different languages (Cutler & Condon, 2023).

***Practical implications***

While there are still limits to what LLMs can do, these models should become increasingly useful as their weaknesses are addressed. Researchers can use them in pilot experiments to pre-test predictions concerning how differently people from various cultures would behave under some circumstances. The ability to emulate cross-cultural differences can also be useful for practitioners, who have the ability to account for psychological factors that might lead to differences in consumer or economic behavior (e.g., Alderotti et al., 2023; Sandy et al., 2013) across the world.

We present more details on future directions in the Supplemental Materials.

## 6. Conclusions

Our study provides preliminary evidence that LLMs can mimic cross-cultural differences in personality when asked to play the role of a person from a specific country. This capability suggests promising potential for using LLMs in *in silico* experiments to simulate cross-cultural psychological phenomena. However, limitations were observed, including an upward bias in scores, reduced variation compared to real-world samples, and lower structural validity of the factors. Nonetheless, LLMs can be a valuable tool for social scientists and practitioners, particularly in aiding pilot and validation studies involving different cultures.

# Supplemental Materials

Large language models can replicate cross-cultural differences in personality

## Part A. Supplementary notes

### Why do large language models replicate cross-cultural differences in personality?

Our analysis suggests that large language models (LLMs) can potentially replicate cross-cultural personality differences – adding to the earlier body of research connecting personality and LLMs (Jiang et al., 2023; Li et al., 2023; Miotto et al., 2022; Pellert et al., 2022; Serapio-García et al., 2023) – but we have not discussed why they may be able to do so.

We presume that this is due to them reflecting underlying differences between people from different countries that are encoded in the text that humans generated (on the Internet, books etc.), which was subsequently used as training data for the LLMs. While traditional questionnaires remain a common approach for measuring personality traits and cross-cultural differences, LLM offer promising new avenues. LLM operate on mechanisms similar to those used to recreate personality structure (Cutler & Condon, 2023). For example, a language model could be prompted to act as an individual from a specific culture and respond to personality-related questions. By generating numerous responses, it could theoretically produce a dataset representing the distribution of given traits in a population. This approach presents intriguing possibilities for personality research, potentially allowing for rapid, large-scale assessments across various cultural contexts.

But there are other possible mechanisms at play, which are discussed below.

#### Reflection of stereotypes

It is well known that machine learning algorithms replicate biases that are present in data that is used to train them. A prominent bias is gender stereotyping, which results, for example, with certain professions being more likely to be associated with men, and others with women (e.g., Bolukbasi et al., 2016; Hendricks et al., 2018; Zhao et al., 2017). What is more relevant is that large language models can also permeate ethnic or cultural stereotypes (Abid et al., 2021). Thus, it is possible that LLMs do not reflect actual differences in personality, but merely stereotypes. If the stereotypes are grounded in real-world characteristics – that is they are somewhat accurate, which has been shown by much research (Jussim et al., 2016) – then the differences in stereotypes that would be captured by LLMs may proxy for actual differences in personality. However, research on stereotype accuracy also suggests that national stereotypes are the type of stereotype that is actually the *least* accurate (Jussim et al., 2016), i.e. the national stereotypes do not correspond to actual cross-cultural differences in personality (Terracciano et al., 2005). Jussim et al. (2016) report that only 43% of correlations between national stereotypes and actual scores on the Big Five exceed a correlation of $r = .30$ based on 7 correlations using representative samples (e.g., Terracciano et al., 2005), and only 17% in 141 correlations exceed this threshold when using convenience samples.

Also, note that personality assessments carried out independently by people from different cultures also show instances of substantial differences in personality, just as it was in the case of Americans and South Koreans that filled out TIPI (in English) and TIPI-K (in Korean), respectively. Arguably, LLMs prompted in Korean are more likely to reflect auto-stereotypes of Koreans about Koreans, but Jussim et al. (2016) argue that these are also inaccurate.

The role that stereotypes play in shaping an LLM's responses could, however, be more systematically investigated. A testable prediction is that the accuracy of LLMs should increase in cases where stereotypes are more accurate. This could include cross-cultural stereotypes on gender differences in personality (Löckenhoff et al., 2014), which are much more accurate than stereotypes on national character (Jussim et al., 2016).

**Imprinting of values via Reinforcement Learning from Human Feedback (RLHF)**

When people interact with large language models, they interact with a "machine" that has gone through a number of stages to become useful to people (Vaswani et al., 2017). Large language models are based on the Transformer architecture, which consists of various stages (see Hussain et al., 2024 for an explanation). One of the important changes that has allowed the widespread adoption of LLMs into society is Reinforcement Learning from Human Feedback (RLHF). This is a process, in which human annotators shape LLMs by, for example, providing examples of desirable feedback to the LLM and rating its output. It is plausible that during this process humans could "imprint" values into the LLM or – conversely – try to combat stereotypes (Abid et al., 2021).

We would argue that if RLHF might try to minimize any tendencies of LLMs to paint some cultures unfavorably, it would happen by inflating the scores of South Koreans relative to US Americans, as high scores on the Big Five are generally considered socially desirable (Musek, 2007; Saucier & Goldberg, 2003). Overall, there was an upward bias in the scores, which could indeed be the result of inflating scores regardless of the country of origin of the simulated person. The upward bias was stronger for South Korean targets than US American targets, which provides additional evidence that RLHF could be at play.

## Directions for future research

**Use of alternative methods to extract data from LLMs.** To validate that LLMs can indeed robustly replicate cross-cultural differences in personality, it would be advisable to use alternative methods. Instead of asking an LLM to generate text (like we did), one could analyze word embeddings (Cutler & Condon, 2023; Hussain et al., 2024). To illustrate, Cutler and Condon (2023) asked the DeBERTa model (also based on the Transformer architecture) to complete queries such as "Those close to me say I have a [MASK][MASK] and [TERM] personality.", where TERM is an adjectives from Saucier and Goldberg (1996), and the model must fill what is in the MASK fields. Models can be prompted using different languages (e.g., English and Korean), which makes it possible to test whether words in different languages (cultures) aggregate in a similar manner, e.g., consistent to a five factor solution (Cutler & Condon, 2023). A similar procedure could also be used to assess the probability that each of the adjectives is used in a particular personality-specific prompt. This could reflect cross-cultural differences in tendencies, for example, higher extraversion in one culture relative to another one.

**Using different personality models and inventories.** To confirm that the patterns we observed using a very short (ten item) personality inventory measuring the Big Five are valid, future research could see whether findings hold using longer inventories, such as the 44-item Big Five Inventory (John et al., 2012) or the 120-item IPIP-NEO-120 inventory (Johnson, 2014). Our recommendation would be for researchers to use longer inventories with stronger properties than the very short TIPI (specifically with structural and metric invariance), which would enable to test whether there is structural and metric invariance when asking the LLM to imagine (1) being a person from different countries using the same inventory, (2) using different versions of the inventory (i.e., adapted to different cultures, and in different languages). An alternative course would be to use other prominent, non-Big Five personality models, such as HEXACO (Lee & Ashton, 2004), which also show cross-cultural differences, although in a somewhat flawed manner, as HEXACO-100 does not show scalar invariance (Thielmann et al., 2020).

**Investigating other patterns present in studies on personality.** There are substantial gender differences in personality, which vary across cultures (Costa Jr. et al., 2001; Giolla & Kajonius, 2019). LLMs could be used to try to replicate other personality patterns. These do not necessarily how to vary cross-culturally, e.g., it is known that neuroticism decreases with age (Roberts & Mroczek, 2008). LLMs could be prompted to play the role of a people of different genders and of different ages, to see whether these changes resulting from age are also replicated.

**Using different LLMs.** We have used GPT from OpenAI (OpenAI, 2023), which is a proprietary model. Programmatic access to it (via an API) is costly, and there is no ability to access its training data and its weights. Others have proposed the use of simpler, non-proprietary LLMs to study personality. A good recommendation (Cutler & Condon, 2023; Hussain et al., 2024) is to use the open-access Hugging Face interface, which enables easy access to a wide variety of LLMs that also rely on the potent Transformer architecture. Many of these models can entirely by controlled by researchers. This is in contrast to GPT, which is a proprietary model that is updated quite frequently, hindering the replicability of findings. Hugging Face provides access to prominent (partially) open-source alternatives to GPT such as LLaMA (Touvron et al., 2023).

**Methodological issues to consider.** There remain some methodological challenges involved with using LLMs for in-silico experiments. Classical statistical tests – such as the two-sample *t*-test that we used to compare humans with the LLM – rely on the assumption that observations within a sample are independent, while data generated by LLMs is always generated by the very same model, even though it may be configured to exhibit as much variation as possible. Even though we attempted to maximize variability of output by using the maximum temperature setting – making output the least deterministic – simulated scores exhibited less variation than in the original TIPI and TIPI-K datasets. This is likely a reflection of a known flaw of LLMs and other neural networks called "mode collapse" (e.g., Hamilton, 2024), where certain output is overrepresented. Future research should be cognizant of this flaw when carrying out in-silica experiments (see Hussain et al. (2024) for other recommendations). Hopefully, advancements in LLMs will allow to more accurately reflect variation in personality across humans.

## Part B. Materials used to elicit personality responses

**English/TIPI**

**System message (pre-prompt)**

You are playing the role of an adult from the United States./

You are playing the role of an adult from South Korea.

**Prompt**

Here are a number of personality traits that may or may not apply to you. Please write, separately for each item and in square brackets, a number that indicates the extent to which you agree or disagree with that statement. You should rate the extent to which the pair of traits applies to you, even if one characteristic applies more strongly than the other.\n1 = Disagree strongly\n2 = Disagree moderately\n3 = Disagree a little\n4 = Neither agree nor disagree\n5 = Agree a little\n6 = Agree moderately\n7 = Agree strongly\n\nI see myself as:\n1. _____ Extraverted, enthusiastic.\n2. _____ Critical, quarrelsome.\n3. _____ Dependable, self-disciplined.\n4. _____ Anxious, easily upset.\n5. _____ Open to new experiences, complex.\n6. _____ Reserved, quiet.\n7. _____ Sympathetic, warm.\n8. _____ Disorganized, careless.\n9. _____ Calm, emotionally stable.\n10. _____ Conventional, uncreative.\nI see myself as:”

**Korean/TIPI-K**

**System message (pre-prompt)**

당신은 미국에서 온 성인 역할을 하고 있습니다./

당신은 한국에서 온 성인 역할을 하고 있습니다.

**Prompt**

다음은 귀하의 성격 특성들에 관한 진술들입니다. 귀하가 각 진술에 동의하거나 동의하지 않는 정도를 각 진술 꺾쇠 괄호에 숫자로 [1 - 7] 표시해 주십시오. 각 진술은 두 가지의 성격 특성들을 표현하고 있습니다. 두 개의 성격 특성들 중 하나의 성격 특성이 다른 하나의 성격 특성보다 귀하의 성격을 더 잘 표현하더라도 그 두 개의 성격 특성들 모두가 귀하에게 적용되는 정도를 아래 숫자들로 평가해 주십시오. 대답을 꺾쇠괄호에서 ’1. [5]’처럼 쓰십시오.\n1 = 전혀 동의하지 않는다\n2 = 동의하지 않는다\n3 = 그다지 동의하지 않는다\n4 = 중간이다\n5 = 어느 정도 동의한다\n6 = 동의한다\n7 = 매우 동의한다\n\n내가 보기에 나 자신은:\n1. _____ 외향적이다. 적극적이다.\n2. _____ 비판적이다. 논쟁을 좋아한다.\n3. _____ 신뢰할수있다. 자기 절제를 잘한다. \n4. _____ 근심 걱정이 많다. 쉽게 흥분한다. \n5. _____ 새로운

경험들에 개방적이다. 복잡다단하다. \n6. _____ 내성적이다. 조용하다.\n7. _____ 동정심이 많다. 다정다감하다. \n8. _____ 정리정돈을 잘못한다. 덤벙댄다. \n9. _____ 차분하다. 감정의 기복이 적다. \n10. _____ 변화를 싫어한다. 창의적이지 못하다. \n내가 보기에 나 자신은:

# Part C. Supplementary analyses

Note that we conducted a similar (but less detailed) analysis on an earlier version of GPT-3.5. However, the resurgence of studies suggesting the vast superiority of GPT-4 relative to GPT-3.5 for some tasks (Bubeck et al., 2023) led us to reexamine this issue, using GPT-4 as the workhorse, and a stronger design for the study. It is worth noting that GPT-4 cost (at the moment we ran our study) roughly 25 times as much to run than GPT-3.5, and thus the costs of running large-sample simulations become non-negligible for GPT-4. We should also add that since the study was conducted, newer versions of GPT have become available. These also have different price tiers: the more advanced model (GPT-4o) is roughly 33 times as expensive as the less expensive model (GPT-4o-mini). Researchers should remain cognizant of these substantial price differences, which will only become more pronounced when using longer inventories.

**Validity of output.** Practically all of the output from GPT-4 was in the correct format, whereas 19% of the output from GPT-3.5 was in the incorrect format (answers were not provided in square brackets, or more than two answers were given to some items). This was essentially due to input being more often in the incorrect format when the input was in Korean (35% of output was in an erroneous format) rather than in English (3% error rate).

**Structural validity.** GPT-4 generated data that had $\alpha$s between 0.07 (Agreeableness) to 0.54 (Extraversion) – using TIPI – and between 0.04 (Agreeableness) to 0.70 (Extraversion) – using TIPI-K. Perhaps the most surprising differences are for Emotional Stability, which had an $\alpha$ of .73 in the original study, but only .19 (TIPI) and .16 (TIPI-K) using the simulated data. However, GPT-3.5 generated data that was entirely inconsistent, with $\alpha$s ranging from -.30 to .07 for TIPI and from -.18 to .00 for TIPI-K. Therefore, we performed our main analysis solely on GPT-4.

**Accuracy.** Given that the output from GPT-3.5 was often in the incorrect format (when prompted in Korean), and the structural validity of the output was inconsistent, comparisons between GPT-4 and GPT-3.5 are problematic. However, it is nonetheless worth noting that GPT-3.5 ($M$ = 1.41) was *more* accurate than GPT-4 ($M$ = 1.65) when prompted in English. In contrast, when prompted in Korean, GPT-4 ($M$ = 1.54) was more accurate than GPT-3.5 ($M$ = 1.64). It is unclear why this pattern emerged (mainly, the slight advantage of GPT-3.5 in English), although one could speculate that this could be due to the intensity of Reinforcement Learning from Human Feedback between these generations, which has become more pronounced in newer generations.

# Part D. Supplementary tables and figures

**Table S1**

*Differences in the Big Five between US and South Korea using the IPIP-NEO-120 (Johnson, 2014; Kajonius & Giolla, 2017)*

| Factor | US | South Korea | Difference |
| --- | --- | --- | --- |
| Extraversion | 0.09 | –0.22 | 0.31 |
| Agreeableness | 0.03 | –0.47 | 0.50 |
| Conscientiousness | 0.20 | –0.13 | 0.33 |
| Neuroticism | 0.06 | 0.10 | –0.04 |
| Openness to Experiences | –0.22 | –0.30 | 0.08 |

*Note*. This table presents standardized mean scores ($z$-scores) from Table 2 in Kajonius and Giolla (2017), showing how the mean for a specific country deviates for the mean score across all countries that were investigated. In Gosling et al. (2003) and in the main text of the manuscript, we report “Emotional Stability” instead of “Neuroticism”.

**Table S2**

*Difference between US and South Korea scores on Big Five using different inventories*

| | **TIPI and TIPI-K** (Gosling et al., 2003; Ha et al., 2013) | **NEO-PI-R** (Piedmont & Chae, 1997) | **NEO-PI-R** (Yoon et al., 2002) | **IPIP-NEO-120** (Kajonius & Giolla, 2017) |
|---|---|---|---|---|
| Extraversion | – | – | – | – |
| Agreeableness | – | – | – | – |
| Conscientiousness | – | – | – | – |
| Neuroticism | + | + | + | n.s. |
| Openness to Experiences | – | – | – | – |

*Note.* + and - indicate that South Koreans had a higher and lower score than US Americans, respectively. In Gosling et al. (2003) and in the main text of the manuscript, we report “Emotional Stability” instead of “Neuroticism”.

**Table S3**

*Reliability metrics for actual and simulated scores*

| | **Cronbach's α** | | | | | **Guttman's $\lambda_6$** | | | |
|---|---|---|---|---|---|---|---|---|---|
| | TIPI | | | TIPI-K | | TIPI | | TIPI-K | |
| | **Original study** | GPT-4 | GPT-3.5 | GPT-4 | GPT-3.5 | GPT-4 | GPT-3.5 | GPT-4 | GPT-3.5 |
| Extraversion | **0.68** | 0.54 | –0.30 | 0.70 | –0.00 | 0.54 | –0.30 | 0.56 | –0.00 |
| Agreeableness | **0.40** | 0.07 | –0.02 | 0.04 | –0.06 | 0.04 | –0.01 | 0.02 | –0.04 |
| Conscientiousness | **0.50** | 0.31 | –0.06 | 0.17 | –0.06 | 0.19 | –0.03 | 0.10 | –0.03 |
| Emotional Stability | **0.73** | 0.19 | 0.02 | 0.16 | –0.18 | 0.10 | 0.01 | 0.09 | –0.08 |
| Openness to Experiences | **0.45** | 0.26 | 0.07 | 0.23 | –0.05 | 0.15 | 0.04 | 0.13 | –0.03 |

*Note.* Computed using R package psych (Revelle, 2023). There is no data for Cronbach's α for TIPI-K (Ha et al., 2013).

**Table S4**

*The results of Spearman-Brown formula for GPT-generated TIPI and TIPI-K scores*

| Inventory | Extraversion | Agreeableness | Conscientiousness | Emotional Stability | Openness to Experiences |
|---|---|---|---|---|---|
| TIPI | .54 | .07 | .31 | .19 | .27 |
| TIPI-K | .72 | .04 | .18 | .17 | .24 |

**Figure S1**

*Mean scores across the Big Five for inventory-target congruent pairs: a comparison of GPT-4 and GPT-3.5*

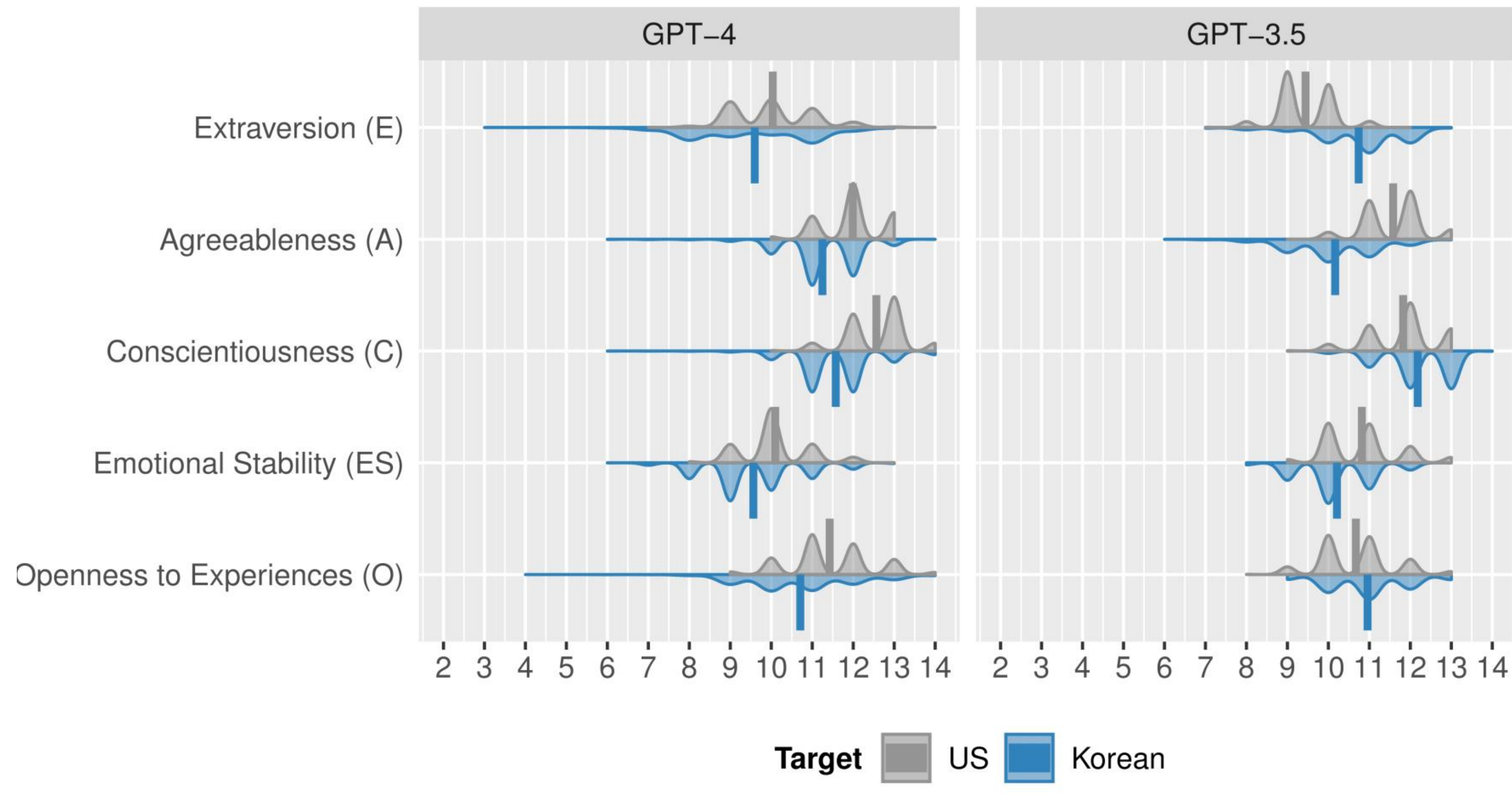